\documentclass[conference]{IEEEtran}
\IEEEoverridecommandlockouts

\usepackage{cite}
\usepackage{amsmath,amssymb,amsfonts}
\usepackage{algorithmic}
\usepackage{graphicx}
\usepackage{textcomp}
\usepackage{xcolor}
\def\BibTeX{{\rm B\kern-.05em{\sc i\kern-.025em b}\kern-.08em
    T\kern-.1667em\lower.7ex\hbox{E}\kern-.125emX}}
    
\usepackage[pagebackref=false,breaklinks=true,colorlinks,citecolor=blue,bookmarks=false]{hyperref}
\usepackage{caption}
\usepackage{booktabs}
\usepackage{pifont}
\usepackage{bm}
\usepackage{multirow}
\usepackage{graphicx}
\usepackage{indentfirst}
\usepackage{lipsum} 
\usepackage{array}    
\usepackage{subcaption} 
\usepackage{makecell}

\title{VAGeo: View-specific Attention for Cross-View Object Geo-Localization}
\author{Zhongyang Li$^{1}$ \hspace{0.1in} Xin Yuan$^{1,2,*}$\thanks{*Corresponding author (yuanxincherry@gmail.com; liuwei@wust.edu.cn).} \hspace{0.1in} Wei Liu$^{1,*}$  \hspace{0.1in} Xin Xu$^{1,2}$\hspace{0.1in}\\
$^{1}$ School of Computer Science and Technology, Wuhan University of Science and Technology\\
$^{2}$ Hubei Province Key Laboratory of Intelligent Information Processing and Real-Time Industrial System
}

\begin{document}

%
%
%

\captionsetup{
  position=top
}
%
\maketitle
\begin{abstract}
Cross-view object geo-localization (CVOGL) aims to locate an object of interest in a captured ground- or drone-view image within the satellite image. However, existing works treat ground-view and drone-view query images equivalently, overlooking their inherent viewpoint discrepancies and the spatial correlation between the query image and the satellite-view reference image. To this end, this paper proposes a novel View-specific Attention Geo-localization method (VAGeo) for accurate CVOGL. Specifically, VAGeo contains two key modules: view-specific positional encoding (VSPE) module and channel-spatial hybrid attention (CSHA) module. In object-level, according to the characteristics of different viewpoints of ground and drone query images, viewpoint-specific positional codings are designed to more accurately identify the click-point object of the query image in the VSPE module. In feature-level, a hybrid attention in the CSHA module is introduced by combining channel attention and spatial attention mechanisms simultaneously for learning discriminative features. Extensive experimental results demonstrate that the proposed VAGeo gains a significant performance improvement, \textit{i.e.}, improving acc@0.25/acc@0.5 on the CVOGL dataset from 45.43\%/42.24\% to 48.21\%/45.22\% for ground-view, and from 61.97\%/57.66\% to 66.19\%/61.87\% for drone-view.

\end{abstract}
\section{Introduction}


The cross-view image geo-localization (CVGL) task is typically investigated as a retrieval problem, where the images captured from a smartphone or drone perspective serve as query images, and satellite images serve as reference images~\cite{rodrigues2023semgeo,li2024hadgeo,sogi2024disaster}. Zhai \textit{et al.}~\cite{5} and Liu \textit{et al.}~\cite{6} introduced the CVUSA and CVACT datasets, respectively, to support research on ground-view missions. Similarly, the University-1652~\cite{zheng2020university} and SUES-200~\cite{zhu2023sues} datasets have made substantial contributions to research involving drone viewpoints. Early methods primarily relied on hand-crafted feature extraction techniques~\cite{2,3,4}. Some metric learning approaches~\cite{compare2, compare3, compare5, direct1, direct2, direct3, workman2015wide, guo2022soft, sun2019geocapsnet, shi2022accurate, yuan2023osap} focused on directly learning viewpoint-invariant features. In contrast, geometry-based methods leverage geometric transformations like polar transformations~\cite{7,8,9,10} to minimize viewpoint discrepancies between ground and satellite images, aligning the content distribution of satellite images with ground panoramas. To further mitigate the differences between perspectives, some works~\cite{11,12,13,14,15,tian2021uav} employed Generative Adversarial Networks (GANs) to generate intermediate query or reference samples, facilitating the localization process.

\begin{figure}[t] 
    \centering
    \includegraphics[width=\linewidth]{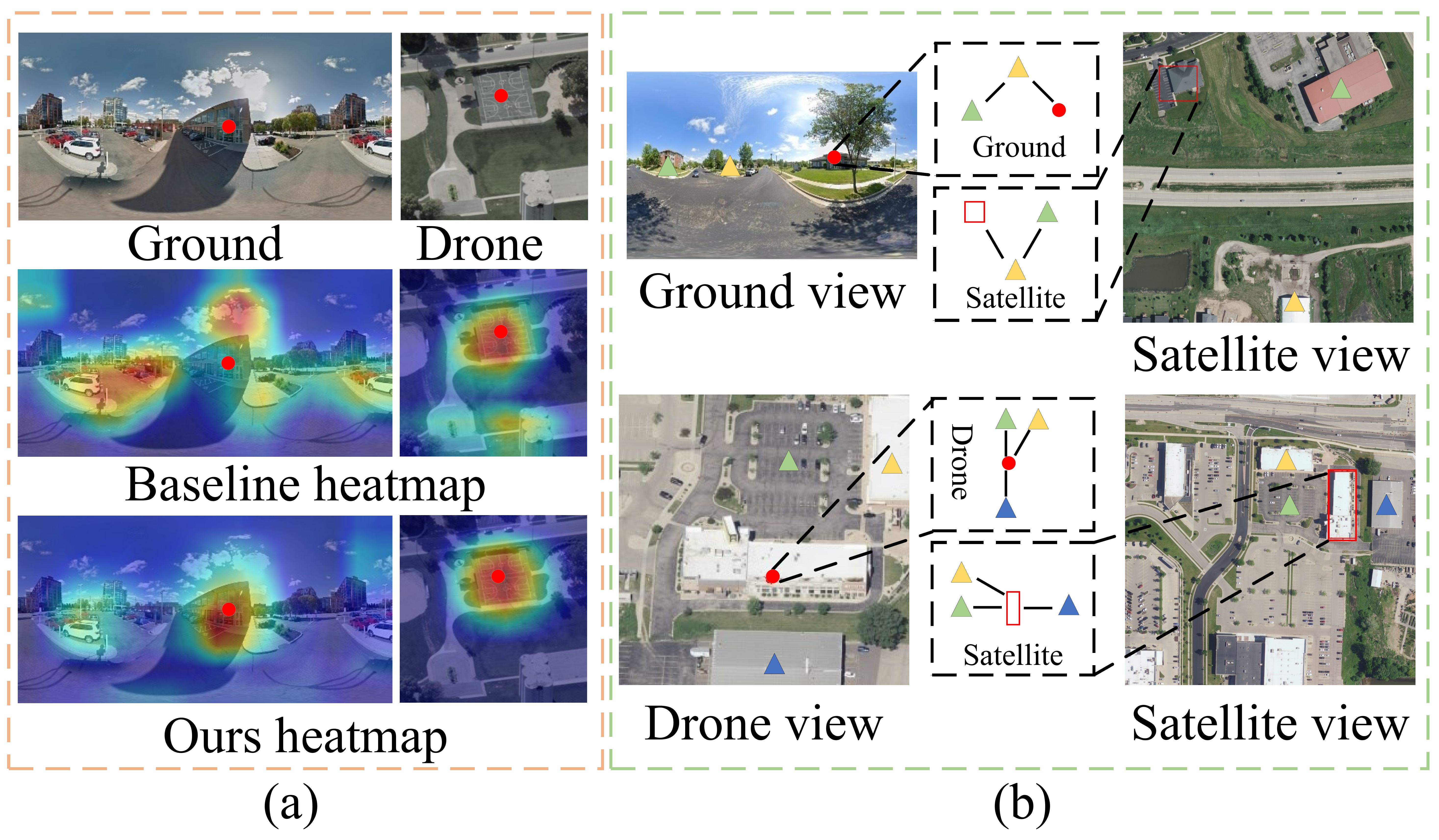}
    \caption{(a) Difference of the activation maps generated by DetGeo~\cite{sun2023cross} and our method from ground- and drone-views. 
    (b) The spatial correlation of ground- and drone-views resembles that of satellite images, allowing the surroundings to serve as a discriminative knowledge. 
    Red click points denote target objects in the query image, red boxes indicate target objects in the reference image, and colored triangles represent potential positive contextual information.
    \textit{Best viewed in color}.
    }
    \label{fig:image_1}
    \vspace{-3mm}
\end{figure}
However, CVGL can only locate the approximate position of the query image but not the location of the specific object. To address this limitation, Sun~\textit{et al.}~\cite{sun2023cross} proposed the cross-view object geo-localization (CVOGL) task, which first employs positional coding to identify the object's location, followed by similarity matching with the local features of satellite images to achieve object-level localization. The technology can also be applied to small object detection\cite{SODetection}. However, using the same positional encoding for both ground and drone views introduces errors due to viewpoint variability and scale inconsistencies, as shown in Fig.~\ref{fig:image_1} (a). Additionally, manually setting the positional encoding to indicate only the approximate object location limits the model’s ability to accurately capture recognizable features. Fig.~\ref{fig:image_1} (b) demonstrates the spatial viewpoint correlation of query and satellite images.

Our work can be summarized as follows: 1) We introduce view-specific positional encoding methods that consider the limited perspectives of ground images and the spatial correlation between drones and satellite images. 2) We employ a combination of channel and spatial attentions to process the features extracted by backbone thus enabling the model to focus on discriminative features. The strategy utilizes the keypoint information to aid the localization.

\begin{figure}[t]
    \centering
    \makebox[\columnwidth][l]{\includegraphics[width=0.95\columnwidth]{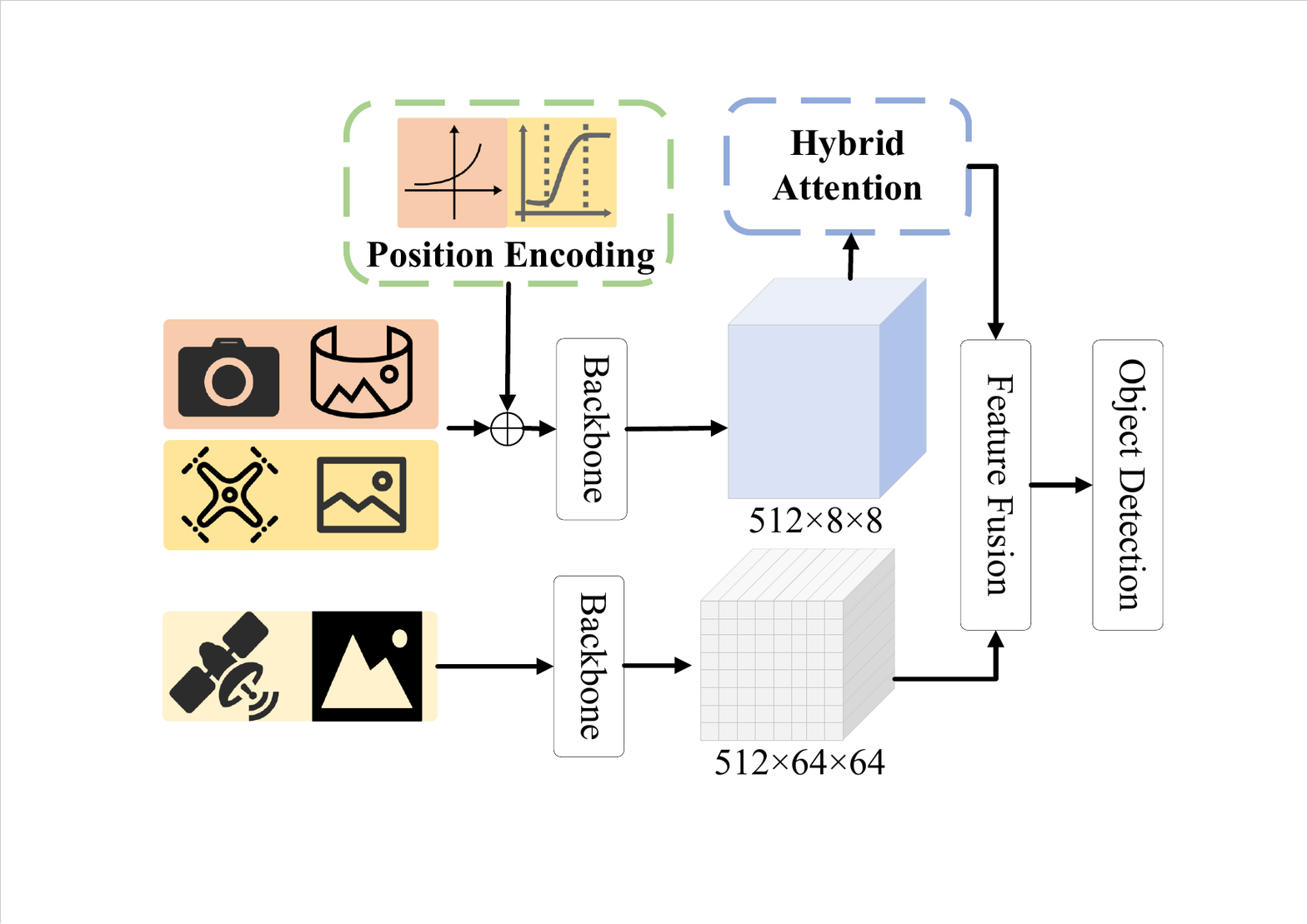}} 
    \caption{Overall architecture of our proposed VAGeo.}
    \label{fig:model}
\end{figure}



\section{Proposed Method}
\label{sec:pagestyle}
The CVOGL task is divided into two steps. The first step directs the model to identify the target object in the query image, while the second step focuses on localizing the target object in the satellite image. We follow \cite{sun2023cross}, which employs a two-branch network to separately extract features from the query image and the reference image. To accommodate the characterization of query views, we employ VSPE to direct the model's attention toward the target object. This is followed by applying CSHA to the query features to emphasize discriminative aspects.  These refined features are then fused with the satellite image features to pinpoint the target object within the satellite image. The model structure is shown in Fig.~\ref{fig:model}.
\begin{table*}[ht]
    \centering
    \caption{Performance comparison with SOTA methods on the CVOGL dataset. Best results are marked in \textbf{bold}.}
    \resizebox{\linewidth}{!}{
    \begin{tabular}{|c|c|c|c|c|c|c|c|c|c|}
        \hline
        \multirow{3}{*}{Method} & \multirow{3}{*}{Venue} & \multicolumn{4}{c|}{\textbf{Ground $\to$ Satellite}} & \multicolumn{4}{c|}{\textbf{Drone $\to$ Satellite}} \\
        \cline{3-10}
         & & \multicolumn{2}{c|}{Test} & \multicolumn{2}{c|}{Validation} & \multicolumn{2}{c|}{Test} & \multicolumn{2}{c|}{Validation} \\
        \cline{3-10}
         & & \multicolumn{1}{c}{@0.25} & @0.5 & \multicolumn{1}{c}{@0.25} & @0.5 & \multicolumn{1}{c}{@0.25} & @0.5 & \multicolumn{1}{c}{@0.25} & @0.5 \\
        \hline
        CVM-Net~\cite{compare1} & CVPR'2018 & \multicolumn{1}{c}{4.73} & 0.51 & \multicolumn{1}{c}{5.09} & 0.87 & \multicolumn{1}{c}{20.14} & 3.29 & \multicolumn{1}{c}{20.04} & 3.47 \\
        Polar-SAFA~\cite{8} & NIPS'2019 & \multicolumn{1}{c}{20.66} & 3.19 & \multicolumn{1}{c}{19.18} & 2.71 & \multicolumn{1}{c}{37.41} & 6.58 & \multicolumn{1}{c}{36.19} & 6.39 \\
        SAFA~\cite{8} & NIPS'2019 & \multicolumn{1}{c}{22.20} & 3.08 & \multicolumn{1}{c}{20.59} & 3.25 & \multicolumn{1}{c}{37.41} & 6.58 & \multicolumn{1}{c}{36.19} & 6.39 \\
        L2LTR~\cite{compare3} & NIPS'2021 & \multicolumn{1}{c}{10.69} & 2.16 & \multicolumn{1}{c}{12.24} & 1.84 & \multicolumn{1}{c}{38.95} & 6.27 & \multicolumn{1}{c}{38.68} & 5.96 \\
        RK-Net~\cite{compare2} & TIP'2022 & \multicolumn{1}{c}{7.40} & 0.82 & \multicolumn{1}{c}{8.67} & 0.98 & \multicolumn{1}{c}{19.22} & 2.67 & \multicolumn{1}{c}{19.94} & 3.03 \\
        TransGeo~\cite{compare5} & CVPR'2022 & \multicolumn{1}{c}{21.17} & 2.88 & \multicolumn{1}{c}{21.67} & 3.25 & \multicolumn{1}{c}{35.05} & 6.37 & \multicolumn{1}{c}{34.78} & 5.42 \\
        DetGeo~\cite{sun2023cross} & TGRS'2023 & \multicolumn{1}{c}{45.43} & 42.24 & \multicolumn{1}{c}{46.70} & 43.99 & \multicolumn{1}{c}{61.97} & 57.66 & \multicolumn{1}{c}{59.81} & 55.15 \\
        FRGeo~\cite{compare6} & AAAI'2024 & \multicolumn{1}{c}{8.12} & 1.31 & \multicolumn{1}{c}{7.80} & 0.87 & \multicolumn{1}{c}{11.41} & 2.67 & \multicolumn{1}{c}{13.22} & 2.06 \\
        GeoDTR+~\cite{compare7} & AAAI'2024 & \multicolumn{1}{c}{14.29} & 5.14 & \multicolumn{1}{c}{14.08} & 1.95 & \multicolumn{1}{c}{16.03} & 4.73 & \multicolumn{1}{c}{15.71} & 3.68 \\
        \hline
        \textbf{VAGeo} & This work & \multicolumn{1}{c}{\textbf{48.21}} & \textbf{45.22} & \multicolumn{1}{c}{\textbf{47.56}} & \textbf{44.42} & \multicolumn{1}{c}{\textbf{66.19}} & \textbf{61.87} & \multicolumn{1}{c}{\textbf{64.25}} & \textbf{59.59} \\
        \hline
    \end{tabular}}
    \label{tab:compare}
\end{table*}

\begin{table}[ht]
    \centering
    \caption{Ablation of positional coding: Drone $\to$ Satellite.}
    \resizebox{\linewidth}{!}{
    \begin{tabular}{|c|cc|cc|}
        \hline
        \multirow{3}{*}{Weight} & \multicolumn{4}{c|}{\textbf{Drone $\to$ Satellite}} \\
        \cline{2-5}
         & \multicolumn{2}{c|}{Test} & \multicolumn{2}{c|}{Validation} \\
        \cline{2-5}
         & @0.25 & @0.5 & @0.25 & @0.5 \\
        \hline
        \textbf{[0.60,0.15,0.15,0.10]} & \textbf{65.26} & \textbf{60.74} & \textbf{62.30} & \textbf{57.75} \\
        \text{[0.60,0.20,0.15,0.05]} & 60.02 & 55.19 & 56.66 & 52.55 \\
        \text{[0.40,0.30,0.20,0.10]} & 59.82 & 55.70 & 57.96 & 54.17 \\
        \text{[0.50,0.30,0.10,0.10]} & 58.27 & 53.96 & 56.45 & 52.00 \\
        \text{[0.70,0.15,0.10,0.05]} & 61.66 & 57.04 & 61.86 & 57.64 \\
        \text{[0.80,0.10,0.05,0.05]} & 61.77 & 57.45 & 62.73 & 58.61 \\			
	\text{[0.60,0.25,0.10,0.05]} & 61.56 & 55.91 & 60.24 & 55.04 
        \\		
        \text{[0.90,0.05,0.05,0.00]} & 63.00 & 59.92 & 64.03 & 59.7 \\
        \hline
    \end{tabular}}
    \label{tab:drone_position}
\end{table}

\subsection{Positional Encoding}
\begin{figure}[t]
    \centering
    \makebox[\columnwidth][l]{\includegraphics[width=1\columnwidth]{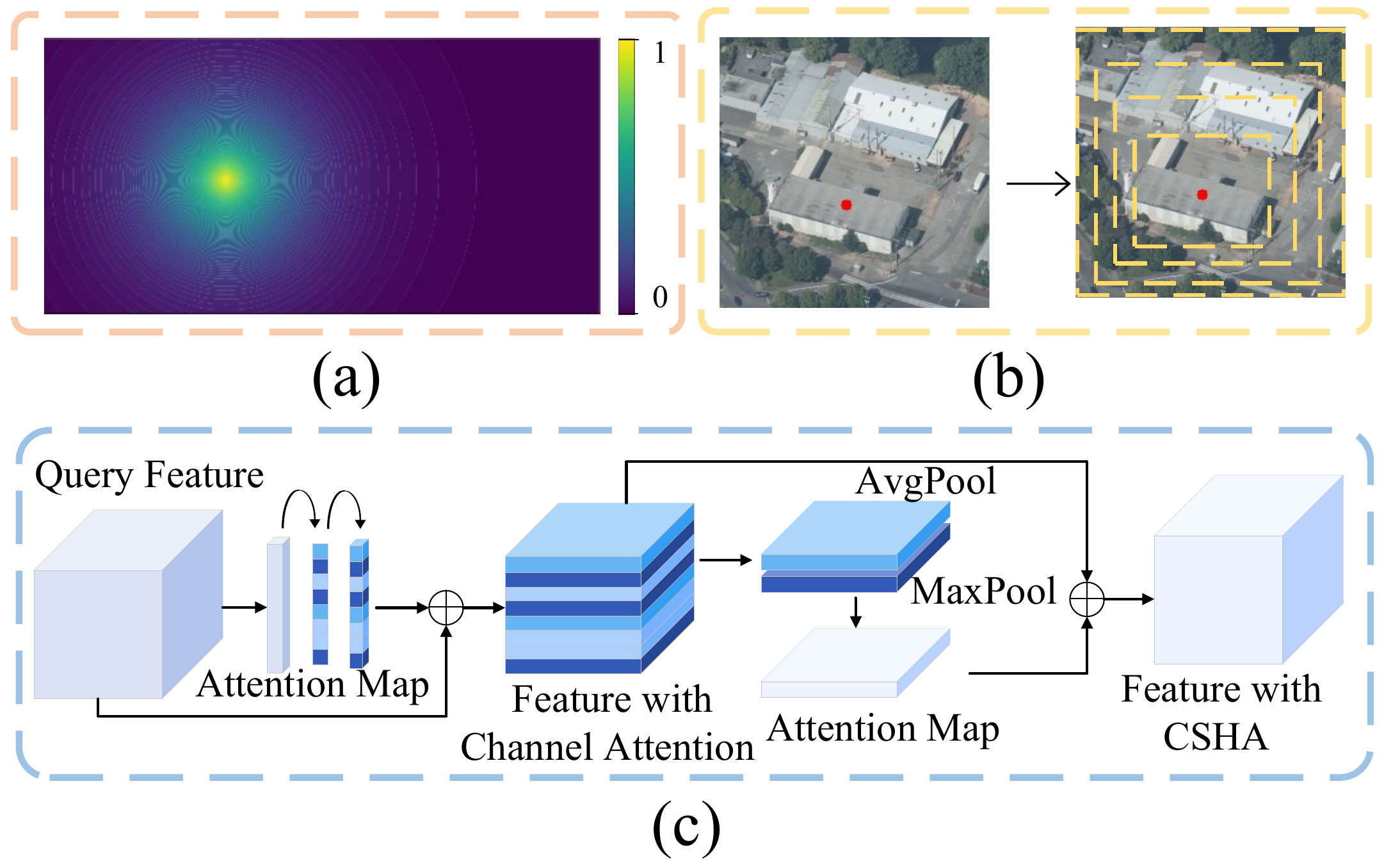}} 
    \caption{The details of our method: (a) VSPE for ground-view, (b) VSPE for drone-view, and (c) CSHA.}
    \label{fig:position_attention}
    \vspace{-2mm}
\end{figure}

\textbf{Ground view}: In this task, the ground query image is a panoramic image that often includes only a partial view of the target object, with the remaining content typically comprising interference elements such as the sky and shadows. These extraneous elements can negatively affect the feature extraction of the target object. Positional encoding is designed to direct the model's attention more towards the target object's features while extracting features from the entire image. We compute the Euclidean distance from each pixel to the click point to form a correlation matrix, then apply a Laplacian distribution, giving greater weight to regions near the click point and reducing interference. We define $\mathbf{P}_k$ to denote the positional-encoded result matrix centered on the click point $\textit{p}_k$ with the same size as the query image. This process is illustrated in Fig.~\ref{fig:position_attention} (a). The formula is as follows:
\begin{align}
\mathbf{P}_k(i,j)=\frac1{2\cdot\mathrm{\sigma}}\cdot exp(\frac{(\|Pixel_k(i,j) - p_k\|_2)^2}{\mathrm{\sigma}})
\end{align}
where $\textit{Pixel}_k(i,j)$ denotes the pixel point position in row $i$ and column $j$, $\|\cdot\|_2$ denotes the Euclidean distance, and $\mathrm{\sigma}$ is the standard deviation used to control the range of the weight decay distribution.

\textbf{Drone view}: Drone and satellite images both capture the top surface and part of the target object's elevation, along with surrounding contextual information. Based on this property, we adaptively partition the image into four square-ring regions centered on the object and assign weights decreasing from near to far, as shown in Fig.~\ref{fig:position_attention} (b).

\subsection{Channel-Spatial Hybrid Attention} 
To fully leverage the extracted contextual information for object localization, we normalize the original features of the query image. Subsequently, channel and spatial attention mechanisms are applied to emphasize key information and highlight target objects within the query image features. Different from existing hybrid attention methods, we explore solutions to the above issues from two aspects: semantic disparities mitigation and multi-semantic guidance. Our channel-spatial hybrid attention (CSHA) is illustrated in Fig.~\ref{fig:position_attention} (c).

\textbf{Channel Attention Mechanism}: 
The channel attention module refines channel features, effectively mitigating semantic disparities and ensuring robust feature integration across channels.
The global information of each channel is extracted through global average pooling, forming a global feature vector. This global perspective helps the model to capture the overall feature distribution. The global feature vector is then nonlinearly transformed using a two-layer fully connected network. The first fully connected layer reduces the feature dimensions, decreases the number of parameters, and introduces nonlinearity. The second fully connected layer restores the feature dimensions to generate the final attention weights. In this way, the model learns complex relationships between different channels. We define the features extracted by backbone as $\mathbf{F^q}$, the channel attention weight map as $\mathbf{X}_\text{channel}$, and the features calculated by channel attention as $\mathbf{F_c^q}\in\mathbb{R}^{\mathrm{B \times C \times H \times W}}$. The computational procedure is defined as follows:
\begin{align}
        \mathbf{X}_\text{channel} &= Sig(\mathbf{W}_2 \cdot \mathrm{ReLU}(\mathbf{W}_1 \cdot \mathbf{\phi(\mathbf{F^q})})) \\
    \mathbf{F_c^q} &= \mathbf{F^q} \times \mathbf{X}_\text{channel} 
\end{align}
where $Sig$ is the Sigmoid activation function, $\phi$ represents the global average pooling operation. $\mathbf{W}_1 \in \mathbb{R}^{C \times \frac{C}{16}}$ and $\mathbf{W}_2 \in \mathbb{R}^{\frac{C}{16} \times C}$ represent the weights of the fully connected layers.

\textbf{Spatial Attention Mechanism}: 
The spatial attention mechanism captures multi-semantic spatial information, ultimately enhancing both global and local feature representations. In our approach, the feature map $\mathbf{F_c^q}$, obtained from the channel attention mechanism, is used as the input to the spatial attention mechanism. The spatial attention weight map $\mathbf{X_{spatial}}$ is then applied to $\mathbf{F_c^q}$, yielding the final feature map $\tilde{F}\in \mathbb{R}^{\mathrm{B\times C\times H\times W}}$.
\begin{align}
\tilde{F} &= \mathbf{F_c^q} \times \mathbf{X}_\text{spatial} \\
    \mathbf{X}_\text{spatial} &= \sigma(\mathrm{ReLU}(\mathrm{BN}(\mathrm{Conv}(\mathbf{P_{cat}})))) \\
    \mathbf{P_{cat}} &= \mathrm{Concat}(\mathbf{P_{avg}}, \mathbf{P_{max}})
\end{align}
where $\mathbf{P}_{\mathrm{avg}}$ represents average pooling, $\mathbf{P}_{\mathrm{max}}$ represents maximum pooling, and $\sigma$ is the Sigmoid activation function.

\begin{table}[ht]
    \centering
    \caption{Ablation of CHSA.}
    \label{tab:combined_att}
    \resizebox{\linewidth}{!}{\begin{tabular}{|c|c|cc|cc|}
        \hline
        \multicolumn{6}{|c|}{\textbf{Ground $\to$ Satellite}} \\
        \hline
        \multirow{2}{*}{\makecell{Channel\\Attention}} & \multirow{2}{*}{\makecell{Spatial\\Attention}} & \multicolumn{2}{c|}{Test} & \multicolumn{2}{c|}{Validation} \\
        \cline{3-6}
         & & @0.25 & @0.5 & @0.25 & @0.5 \\
        \hline
        \ding{55} & \ding{55} & 48.09 & 43.88 & 44.42 & 41.38 \\
        \ding{51} & \ding{55} & 46.04 & 42.14 & 44.42 & 41.17 \\
        \ding{55} & \ding{51} & 47.07 & 42.86 & 46.8 & 43.45 \\
        \ding{51} & \ding{51} & \textbf{48.21} & \textbf{45.22} & \textbf{47.56} & \textbf{44.42} \\
        \hline
        \multicolumn{6}{|c|}{\textbf{Drone $\to$ Satellite}} \\
        \hline
        \multirow{2}{*}{\makecell{Channel\\Attention}} & \multirow{2}{*}{\makecell{Spatial\\Attention}} & \multicolumn{2}{c|}{Test} & \multicolumn{2}{c|}{Validation} \\
        \cline{3-6}
         & & @0.25 & @0.5 & @0.25 & @0.5 \\
        \hline
        \ding{55} & \ding{55} & 65.26 & 60.74 & 62.3 & 57.75 \\
        \ding{51} & \ding{55} & 65.67 & 60.84 & 63.49 & 59.05 \\
        \ding{55} & \ding{51} & 63.93 & 58.79 & 63.71 & 58.18 \\
        \ding{51} & \ding{51} & \textbf{66.19} & \textbf{61.87} & \textbf{64.25} & \textbf{59.59} \\
        \hline
    \end{tabular}}
\end{table}

\section{Experiments}

\textbf{Dataset}: 
We evaluate our method using the CVOGL standard benchmark, which contains 5,836 high-resolution satellite images ($1024 \times 1024$ pixels), 5,279 ground view images($256 \times 512$ pixels), and 5,279 drone aerial views ($256 \times 256$ pixels). The target objects are marketed in the query images (ground and drone images) by click points and in the reference image (satellite image) by bounding boxes (bboxes).

\textbf{Evaluation Settings}: The Intersection over Union (IoU) metric is used to evaluate model performance. Localization accuracy is assessed by calculating the ratio of the intersection to the union of the two bboxes.

\textbf{Implementation Details}: We construct a two-branch network with pre-trained resnet-18 and pre-trained darknet-53, which are used as feature extraction networks for query image and reference image, respectively. The learning rate is initialized to 0.0001 and decays by half every 10 epochs, the batch size is set to 12 and a total of 25 epochs are trained.

\subsection{Comparison with Other Methods}
Since the existing CVGL research is a retrieval task, we follow the Baseline~\cite{sun2023cross} approach to divide the reference satellite image into patches to do similarity comparison with the corresponding query image. The top five patches with the highest similarity scores are selected as the predicted bounding boxes, and we compute the IoU with the ground truth bounding box to evaluate performance.

According to Table~\ref{tab:compare}, in the ground-to-satellite mission, the test performance of our model improves from the best result A of 45.43\% to 48.21\% at an Iou threshold of 25\%, and from 42.24\% to 45.22\% at an IOU threshold of 50\%. In the drone-to-satellite task, the test performance of our model compares to the best score DetGeo in the Iou threshold of 25\% improves the performance by 4.22\% and at Iou threshold of 50\% improves the performance by 4.21\%.
\subsection{Ablation Studies}
In this chapter, we perform parameter ablation for positional coding in different perspectives and modular ablation for hybrid attention mechanisms.

\begin{figure}[t]
    \centering
    \includegraphics[width=\linewidth]{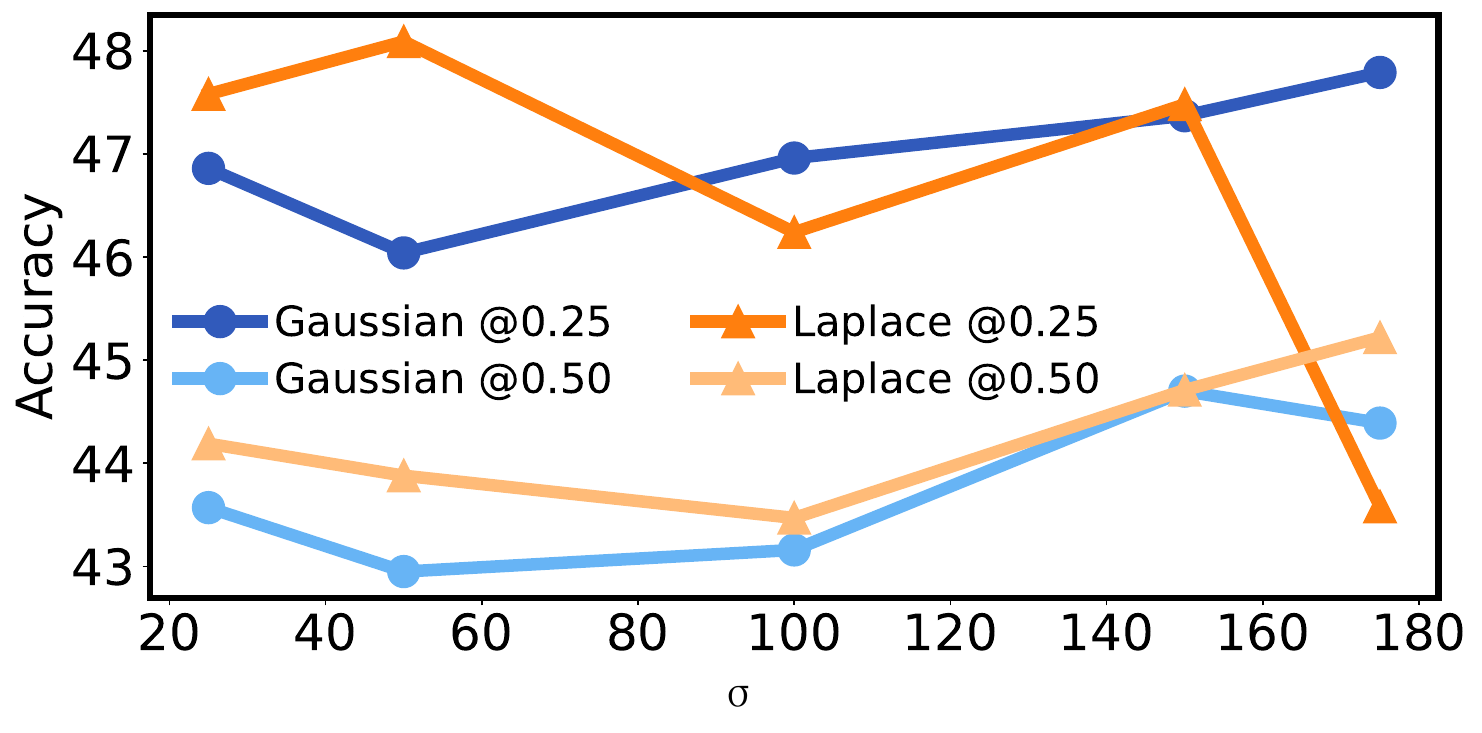}
    \caption{Ablation study of $\sigma$ in VSPE for ground-view.}
    \label{fig:Gposition}
    \vspace{-3mm}
\end{figure}

\textbf{Ablation sduty for VSPE}: Ground-view images offer only partial information about the object's façade and are frequently obscured by distracting elements due to viewing angle limitations. To address this issue, we utilize Gaussian and Laplace distributions with varying standard deviation parameters ($\sigma$) for encoding the ground images. This method enables us to assess how effectively the model focuses on the object. Our experiments reveal that the Laplace distribution, with $\sigma$ set to 25, is more suitable for ground images. The detailed results are presented in Fig.~\ref{fig:Gposition}.


For drone-view images, we perform adaptive region partitioning to adaptively form ring regions with different weights from near to far centered on the target object due to its spatial correlation with satellite images. After the weighting test, we used a weight distribution with inside-out as [0.60,0.15,0.15,0.10] as the positional encoding of the drone viewpoints. The results of the weighting test experiment are shown in Table~\ref{tab:drone_position}.

\textbf{Ablation sduty for CSHA}: We employ a CSHA mechanism to enable the model to learn both target object features and recognizable contextual features surrounding the object. Table~\ref{tab:combined_att} compares the performance of channel attention, spatial attention, and CSHA. The experimental results demonstrate that our CSHA mechanism is the most effective for this task, resulting in more accurate localization of the object.


\begin{figure}[t] 
    \centering
    \includegraphics[width=\linewidth]{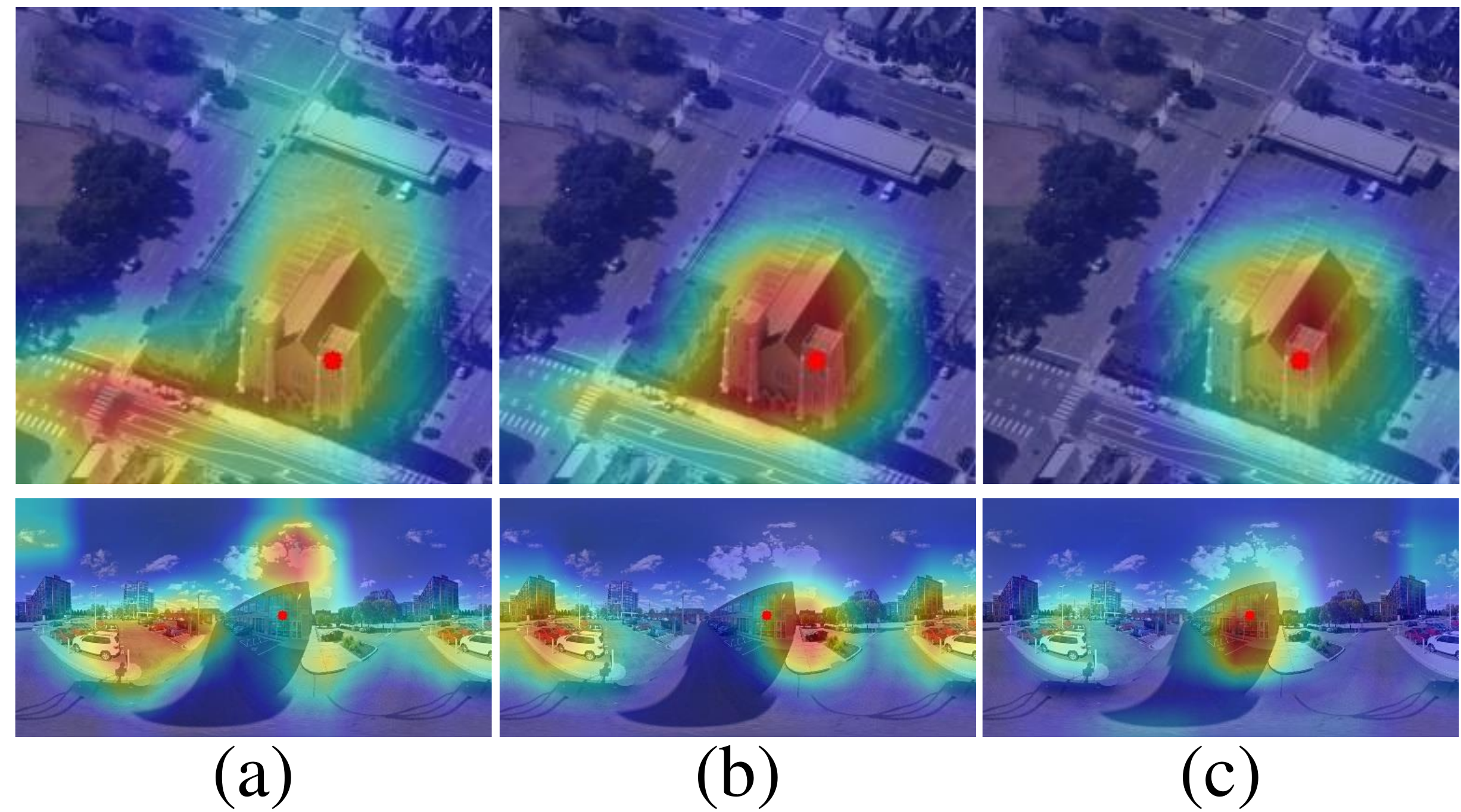} 
    \vspace{1mm}
    \caption{Visualization of heatmaps for ground- and drone- views. (a) Baseline heatmap. (b) Ours with VSPE heatmap. (c) Ours with VSPE and CSHA heatmap.}
    \label{fig:heatmap}
    \vspace{-6mm}
\end{figure}

\begin{figure}[t] 
    \centering
    \includegraphics[width=\linewidth]{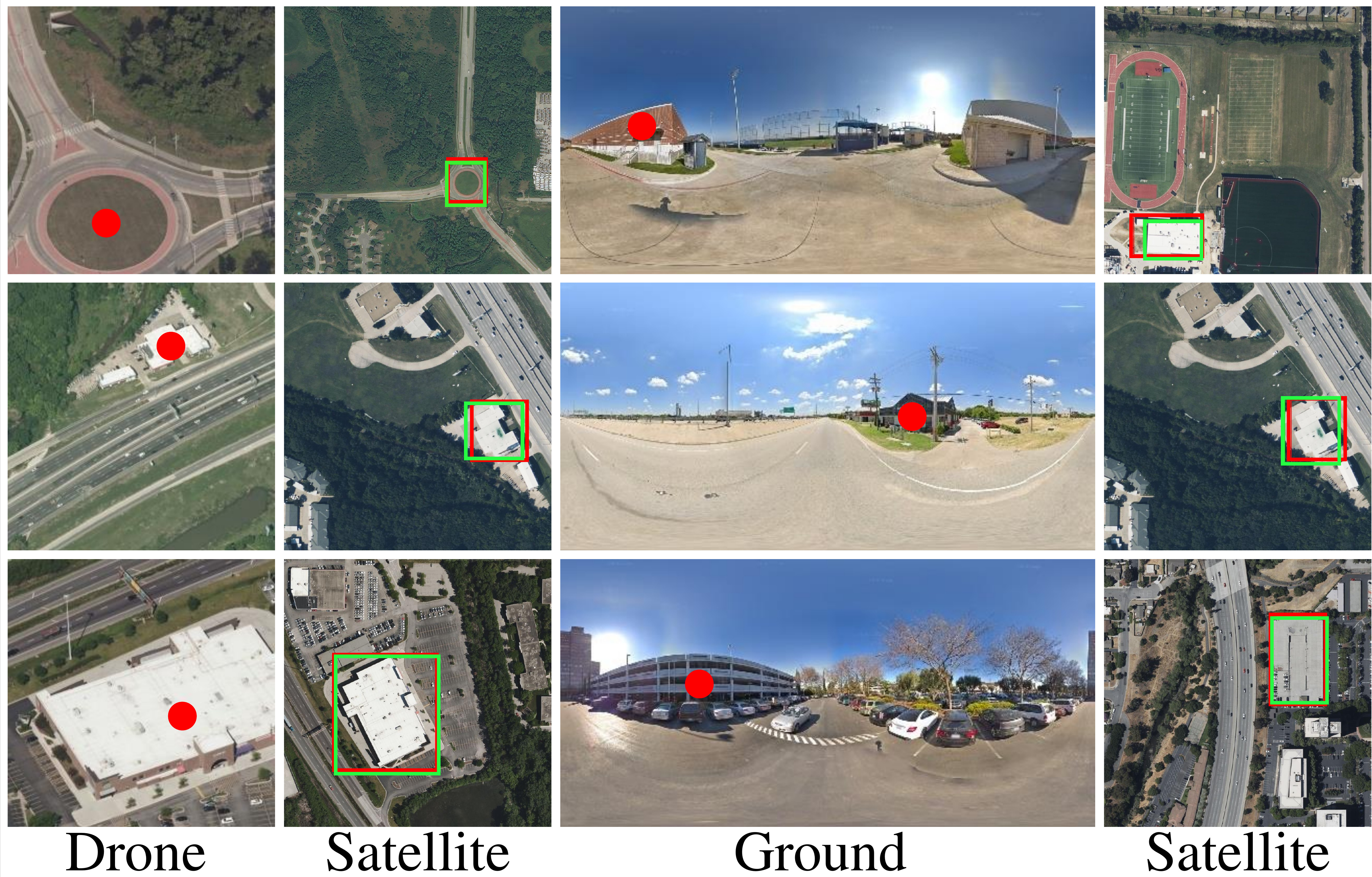} 
    \vspace{1mm}
    \caption{Visualization results of our VAGeo. The red and green bboxes are the predicted and ground truth results, respectively.}
    \label{fig:visual}
    \vspace{-6mm}
\end{figure}

\subsection{Visual Analysis }
Fig.~\ref{fig:heatmap} clearly shows that the VSPE reduces the model's sensitivity to surrounding interference, and with the addition of CSHA, the model further accurately focuses its attention on the target region. Fig.~\ref{fig:visual} visualizes the result of our proposed VAGeo model on the CVOGL task. The close alignment between the green and red bounding boxes highlights the accurate localization achieved by VAGeo.

\section{Conclusion}
In this paper, we proposed VAGeo, a novel approach for CVOGL. We introduced VSPE at the object level, which is tailored to the viewpoint-specific characteristics of query images, effectively addressing challenges related to scale and viewpoint variability. Additionally, we incorporated CSHA at the feature level to enhance feature processing. Our experimental results demonstrate that VAGeo, through the VSPE, enables more accurate focus on the target object, while the CSHA allows the model to autonomously learn distinctive feature dimensions. This combination significantly improves localization accuracy in CVOGL.

\small
\noindent \textbf{Acknowledgments:}
This work was supported by the National Nature Science Foundation of China (No. 62376201). This research was financially supported by funds from Hubei Province Key Laboratory of Intelligent Information Processing and Real-time Industrial System (Wuhan University of Science and Technology) (No. ZNXX2023QNO3), Fund of Hubei Key Laboratory of Inland Shipping Technology and Innovation (NO. NHHY2023004), Key Laboratory of Social Computing and Cognitive Intelligence (Dalian University of Technology), Ministry of Education (No. SCCI2024YB02), and Entrepreneurship Fund for Graduate Students of Wuhan University of Science and Technology (No. JCX2022031, JCX2023049, JCX2023160). We thank all reviewers for their comments.


\bibliographystyle{IEEEtran}
\bibliography{refs}

\end{document}